%% file: root.tex
\documentclass[letterpaper, 10 pt, conference]{ieeeconf}  

\IEEEoverridecommandlockouts                              

\usepackage{amsmath,amsfonts}
\usepackage{algorithmic}
\usepackage{float}
\usepackage{bm}
\usepackage[]{caption}
\usepackage[linesnumbered,ruled,vlined]{algorithm2e}
\usepackage{amsmath, amssymb}
\usepackage{array}
\usepackage[caption=false,font=normalsize,labelfont=sf,textfont=sf]{subfig}
\usepackage{textcomp}
\usepackage{stfloats}
\usepackage{url}
\usepackage{verbatim}
\usepackage{graphicx}
\usepackage{cite}
\usepackage{xcolor}
\usepackage{lipsum}
\usepackage{comment}

\usepackage{tikz}
\usepackage{eso-pic}
\AddToShipoutPictureBG*{%
  \AtPageUpperLeft{%
    \raisebox{-1.75cm}{\makebox[\paperwidth]{%
      \begin{tikzpicture}
        \node[text=black!60, align=center] at (0,0) 
        {This paper has been accepted for publication at the \\
        IEEE/RSJ International Conference on Intelligent Robots and Systems (IROS) 2025};
      \end{tikzpicture}%
    }}%
  }
}

\title{\LARGE \bf
JENGA: Object selection and pose estimation\\ for robotic grasping from a stack
}

\author{Sai Srinivas Jeevanandam$^1$, Sandeep Inuganti$^{1,2}$, Shreedhar Govil$^1$, Didier Stricker$^{1,2}$, Jason Rambach$^1$ \\
$^1$\textit{German Research Center for Artificial Intelligence (DFKI)}, $^2$\textit{RPTU Kaiserslautern} 
    \thanks{Contact:\{sai\_srinivas.jeevanandam, jason.rambach\}@dfki.de}}

\begin{document}

\maketitle

\thispagestyle{empty}
\pagestyle{empty}

\begin{abstract}

Vision-based robotic object grasping is typically investigated in the context of isolated objects or unstructured object sets in bin picking scenarios. However, there are several settings, such as construction or warehouse automation, where a robot needs to interact with a structured object formation such as a stack. In this context, we define the problem of selecting suitable objects for grasping along with estimating an accurate 6DoF pose of these objects. To address this problem, we propose a camera-IMU based approach that prioritizes unobstructed objects on the higher layers of stacks and introduce a dataset for benchmarking and evaluation, along with a suitable evaluation metric that combines object selection with pose accuracy. Experimental results show that although our method can perform quite well, this is a challenging problem if a completely error-free solution is needed. Finally, we show results from the deployment of our method for a brick-picking application in a construction scenario.    

\end{abstract}


\section{Introduction}

In modern robotic applications, such as warehouse automation and domestic assistance, robots frequently encounter scenes where objects are \textit{stacked} rather than merely scattered or isolated. Stacked arrangements introduce substantial \textit{occlusion} and \textit{partial observability}, often leading to ambiguous or erroneous pose estimates. Although a state-of-the-art 6DoF (6 Degrees of Freedom) pose estimation algorithm~\cite{Lin_2024_CVPR} can generate poses for multiple objects, many of these might be \textit{unreliable} or \textit{physically obstructed}, making them unsuitable for grasping. Specifically, objects may be nestled beneath or alongside other objects in the stack, creating challenges that most pose estimation methods do not adequately address.

\begin{figure}[!tbp]
  \centering
  \includegraphics[width=0.48\textwidth]{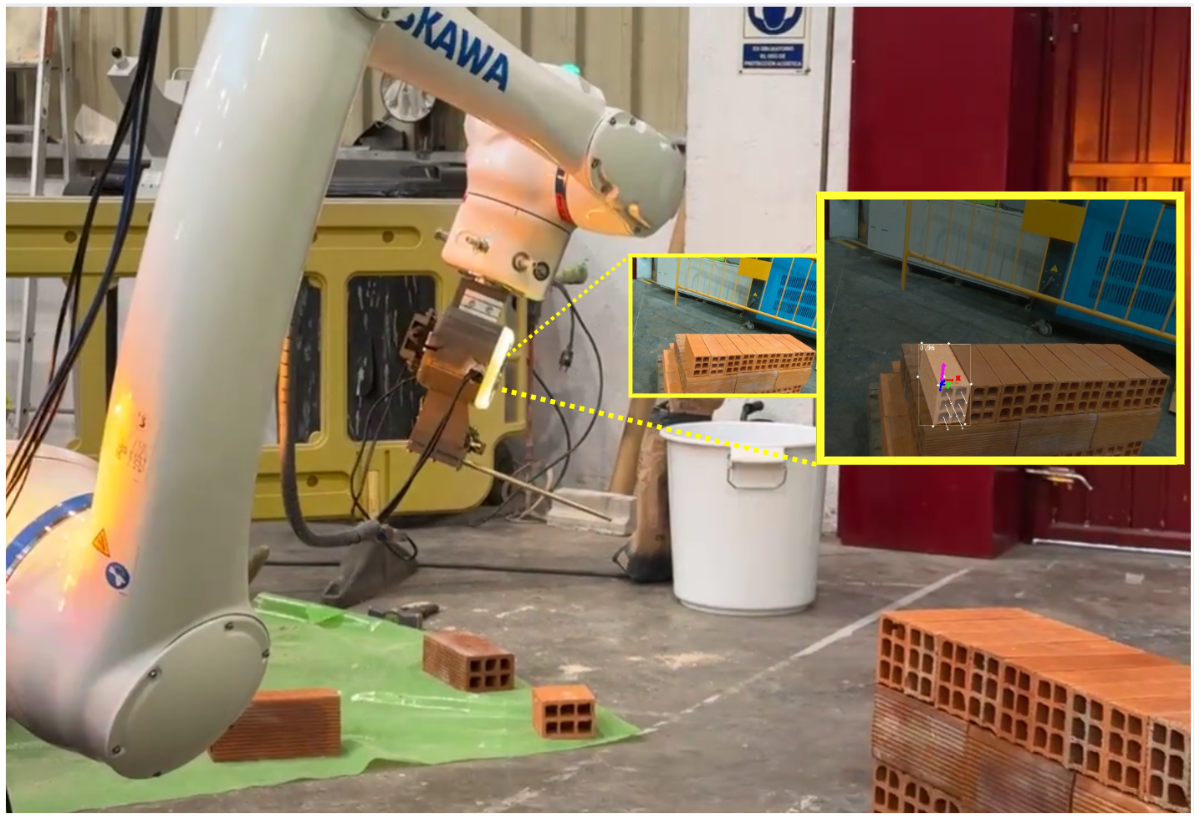}
  \caption{When performing robotic grasping from an object stack, detection and 6DoF object pose are not sufficient. We present an approach that is able to select suitable objects for grasping.}
  \label{fig:teaser}
\end{figure}

On the other hand, research on robotic grasping has put significant effort on the bin-picking scenario where objects are placed in an unordered manner inside a container and are grasped from above~\cite{cordeiro2022bin,drost2017introducing}. In this case object grasping is less challenging and has less potential impact on a failed grasp than picking from a stack. Since object stacks are common structures encountered in daily life, we advocate the need for the development of robotic grasping policies for such structures as well.

To achieve this, we initially created a synthetic dataset of stacked objects (bricks in our case) representing a common use-case of objects stacked in pallets in construction. Further, we formally define the problem of object picking from stacks and design appropriate metrics for its evaluation that combine object detection and pose estimation with object suitability for grasping. We investigate and evaluate baseline approaches for the object selection, such as relying on the confidence of the object detector to select objects of lower occlusion or on inertial sensor measurements to locate the objects on the top of the stack. Finally, we propose a novel method to select suitable grasping candidates that combines pose information, object mask visibility, and inertial sensor cues. In summary, this work introduces the following contributions: 

\begin{itemize}
    \item We formally define the problem of object grasping from structured stacks as opposed to the common bin-picking scenario.
    \item We introduce a novel, procedurally generated dataset for the stacked object pose estimation and grasping task along with a suitable evaluation protocol to quantitatively benchmark solutions.
    \item We propose a novel visual-inertial method for stacked object selection and pose estimation and an evaluation against appropriate baselines.
\end{itemize}

\input{RelatedWork}

\input{problem_formulation}

\input{Methodology}

\input{Dataset}

\input{Experiments}

\section{Conclusion}
In this paper, we formally introduced the problem of selecting an object to grasp from a stack and estimating its pose. Due to the novelty of the problem, it was necessary to develop a new dataset with ground truth object pose and graspability labels to serve as a benchmark for evaluations. Our proposed approach combines visual and inertial information to favor objects with low occlusion that are also situated on the higher levels of the stacks to minimize grasping risk. The quantitative evaluation shows that the proposed method is highly effective compared to baseline approaches, while also confirming that the introduced problem is complex and worthy of additional future research attention due to its significance for real-world robotic application tasks. This is also demonstrated in the deployment of the method on a robotic arm in a construction-domain brick grasping scenario.  
\newpage










\bibliographystyle{IEEEtran}
\bibliography{ref}

\end{document}

%% file: RelatedWork.tex
\section{Related Work}\label{sec:rel_work}
\subsection{Object Pose Estimation}
Object pose estimation plays a crucial role in robotic grasping, enabling robots to localize and manipulate objects in various environments accurately. Since robotic grasping is a task that requires highly accurate poses, we focus on instance-level pose estimation methods where a 3D reference model of the exact object whose pose we wish to estimate, is available. Traditional object pose estimation methods such as Iterative Closest Point (ICP)~\cite{121791} have been widely adopted for aligning known 3D models to observed point clouds or image features. However, these methods often struggle in cluttered or occluded environments. Recently, deep learning approaches have significantly improved object pose estimation by leveraging large datasets. While some early methods directly regressed object poses or 3D bounding boxes~\cite{rad2017bb8,su2021synpo}, methods such as PoseCNN~\cite{XiangSNF18}, and HiPose~\cite{Lin_2024_CVPR} utilize deep feature representations to improve robustness to occlusions and lighting variations. ZebraPose~\cite{Su_2022_CVPR} uses a binary surface encoding representation to establish image-to-model (2D-3D) correspondences leading to a high accuracy method while being computationally efficient and suitable for real-time applications. A complete overview of the state-of-the-art in the topic can be found in the report of the BOP challenge~\cite{hodan2024bop}.

\subsection{Robotic Grasping}

Robotic grasping has been extensively studied in diverse settings—such as picking objects from tables, bins, and conveyor belts. In these scenarios, robots employ a combination of grasping and localization techniques to manipulate or retrieve objects.

Robot grasping methods can be broadly classified into two categories: geometric-based and data-driven~\cite{bohgsurvey, SAHBANI2012326}. Geometric (or sometimes called analytical) methods typically first analyze the shape of the target objects to plan an appropriate grasping pose.
Meanwhile, data-driven (or sometimes called empirical) methods are based on machine learning and have become increasingly popular. They have made substantial improvement in recent years owing to the increased data availability, better computational resources, and algorithmic improvements. We refer the reader to standard surveys on the subject for a more complete treatment~\cite{bohgsurvey, zhang2022roboticgraspingclassicalmodern, learingbasedsurvey}.


Despite substantial progress in various grasping methods, the challenge of handling objects in stacks remains under-explored. Stacked objects introduce unique difficulties, such as ensuring stability and executing precise grasps without disturbing neighboring items. In this paper, we tackle these challenges by proposing a novel camera-IMU based approach and presenting an evaluation benchmark dataset.

%% file: problem_formulation.tex
\section{Problem Formulation}\label{sec:formulation}

\subsection{Notation}
We define a \emph{6DoF pose} as a transformation $\mathbf{T}_{co} = [\mathbf{R}_{co} \in \mathbb{R}^{3 \times 3}, \mathbf{o}_{c} \in \mathbb{R}^{3}]$ between the camera coordinate system $C$ and an object coordinate system $O$.

An inertial sensor (IMU) measures linear acceleration $\mathbf{\alpha} \in \mathbb{R}^3$ and angular velocity $\mathbf{\omega} \in \mathbb{R}^3$ with respect to the sensor coordinate system $S$. When no linear acceleration due to sensor motion is present, the measured acceleration corresponds to the gravity vector $\mathbf{g}_{global} = [0,0,-9.81]$ projected on the sensor coordinate system as $\mathbf{\alpha_{s}}=\mathbf{R}_{sg}\,\mathbf{g}_{global}$, where $\mathbf{R}_{sg} \in \mathbb{R}^{3 \times 3}$ is the rotation matrix from the global coordinate frame $G$ to the sensor coordinate frame $S$.

The rigid transformation between camera and inertial sensor coordinate systems, is given by $\mathbf{T}_{sc} = [\mathbf{R}_{sc} , \mathbf{c}_{s}]$.

\begin{figure}[!htbp]
  \centering
  \includegraphics[width=3in, clip, height=4in, keepaspectratio]{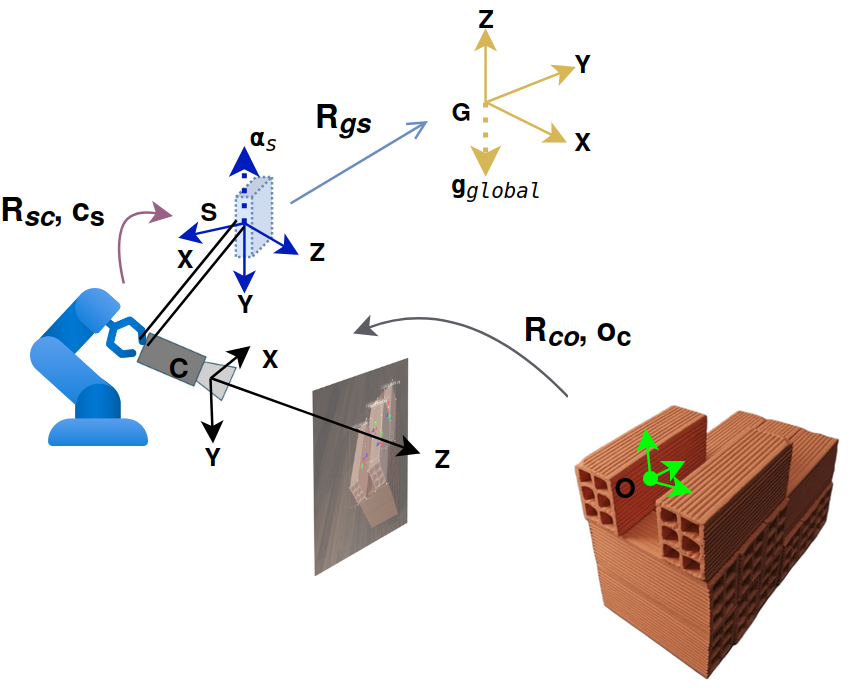}
  \caption{Illustration of the coordinate systems in our pipeline and their transformations}
  \label{fig:problem_formulation}
\end{figure}

\vspace{-1em}
\subsection{Problem Definition}
Given an RGB image $\mathcal{I}$ containing a set of $N$ stacked objects $\mathcal{O}= [O_{1}, \cdots, O_{N}]$ and optionally a corresponding inertial acceleration measurement $\mathbf{\alpha}_{s}^{\mathcal{I}}$, the goal is to identify and estimate the 6DoF poses of the objects that are graspable, both in terms of occlusion and stability of the stack.

Formally, given a set of pose hypotheses

\begin{equation}
\label{eq:poses_set}
\mathbf{\Pi} = \{\mathbf{T}^{1}_{co}, \mathbf{T}^{2}_{co}, \dots, \mathbf{T}^{n}_{co}\},
\end{equation}

we aim to \textit{filter} the raw set of pose estimations $\mathbf{\Pi}$ to yield a smaller, \textit{physically consistent} set of \textbf{graspable} poses:

\begin{equation}
\mathcal{G} = \mathcal{F}(\mathbf{\Pi}, \mathcal{I}, \mathcal{M}, \mathbf{\alpha_{s}}),
\label{eq:filter_function}
\end{equation}

where $\mathcal{M}$ represents 3D object models. 

\noindent Figure \ref{fig:problem_formulation} illustrates the different coordinate frames and their transformations including the pose between the camera and the object.

%% file: Methodology.tex
\section{Methodology}\label{sec:methodology}

\subsection{Overview}
Our approach consists of four major components: \textbf{(i)} detection of 2D regions of interest (RoIs), \textbf{(ii)} 6DoF object pose estimation, \textbf{(iii)} Vision-based object filtering, and \textbf{(iv)} IMU-based object height filtering. This pipeline refines the raw pose hypotheses from \textbf{(i)-(ii)} into a final set of \textit{unobstructed} and \textit{visually consistent} grasping candidates. The modules are detailed below:

\subsection{2D Detection}
An RoI defines the subregion of an image that is hypothesized to contain an object. To achieve this, we use YOLOv8 \cite{yolov8} as the 2D detector. For an input image $\mathcal{I}$, the detector yields a set of bounding boxes \(\mathcal{D}\) for each recognized object. These bounding boxes provide the RoIs needed as input to the pose estimation algorithm.

\subsection{6DoF Object Pose Estimation}
From each bounding box \(d \in \mathcal{D}\), we crop or resize the RoI and feed it into a state-of-the-art pose estimation algorithm. In this work we use the RGB-only variant of \textbf{ZebraPose} by Su et al~\cite{Su_2022_CVPR} due to its accuracy (despite using just an RGB input), computational efficiency(see BOP challenge results~\cite{hodan2024bop}) and its design to handle \textit{challenging viewpoints} and \textit{occlusions}. ZebraPose encodes every point on a 3D model with a binary code, then trains a network to predict those codes for each object pixel in an RGB image, before solving a PnP problem using 2D-3D correspondences to recover the pose. 
The output of each bounding box is produced as a \textbf{6DoF pose} $\mathbf{T}_{co}^{i}$, forming the set of pose estimations $\mathbf{\Pi}$ as described in Eq. \eqref{eq:poses_set}. 

\subsection{Vision-based Object Filtering}
\label{sec:vision-filter}
In the vision-based filtering we adopt a \emph{mask-based} strategy to assess how much of the object is actually visible. As there is a strong correlation between object visibility and graspability, eliminating predicted pose estimates with low visibility is expected to lead to an improvement of our set of grasping candidates.
For each pose $\mathbf{T}_{co}^{i} \in \mathbf{\Pi}$ we use the:
\begin{enumerate}
    \item \textbf{Amodal Mask ($\mathbf{M}_a^i$):} We render the full silhouette of the object (including occluded parts) from the known 3D model $\mathcal{M}$
          using pose $\mathbf{T}_{co}^{i}$.
    \item \textbf{Modal Mask ($\mathbf{M}_m^i$):} We obtain the \emph{visible} mask from the pose estimator's segmentation output.
\end{enumerate}
Using the amodal and modal masks we can define a \textbf{Visibility ratio} as:
\begin{equation}
r^i = \frac{\lvert \mathbf{M}_{m}^i\rvert}{\lvert \mathbf{M}_{a}^i\rvert},    
\end{equation}
where $\lvert \cdot \rvert$ denotes the pixel area. A higher $r^i$ indicates the object is more exposed and thus more \emph{graspable}. If $r^i$ falls below a threshold $\epsilon_{\mathrm{vis}}$, we discard $\mathbf{T}^i$ as insufficiently visible. The remaining set $\mathcal{G}_{v} \subseteq \mathbf{\Pi}$ which contains the poses that pass the visibility threshold is given as

\begin{equation}
\begin{aligned}
\mathcal{G}_v \;&=\; \mathcal{F}_v(\mathbf{\Pi}, \mathcal{I}, \mathcal{M})
\;\\ &= \Bigl\{\,\mathbf{T}^i \in \mathbf{\Pi} \;\Big|\; r^i \;\ge\; \epsilon_{\mathrm{vis}} \Bigr\}.
\end{aligned}
\end{equation}

\subsection{IMU-based height filtering}
\label{sec:height-filter}
The object positions are described by the pose estimator in the camera coordinate systems, i.e. for object $i$ we have $\mathbf{o}_c^i$, which describes the position of the object coordinate origin in the camera coordinate system. However, using the IMU information it is possible to recover $\mathbf{o}_g^{i,z}$, i.e. the position of the object in the global frame with respect to the direction of the gravity vector. To retrieve this, first the object position is transformed to the sensor coordinate system as:
\begin{equation} \label{eq:oc to os}
    \mathbf{o}_s^i = \mathbf{T}_{sc}^{i} \mathbf{o}_{c}^{i}.
\end{equation}
The object position in the global frame $\mathbf{o}_{g}$ would be given as:
\begin{equation}\label{eq:os to og}
    \mathbf{o}_g^i = \mathbf{T}_{gs}^{i} \mathbf{o}_{s}^{i}.
\end{equation}
However, since we are only interested in the position along the gravity axis (height of object position), Eq.\eqref{eq:os to og} is simplified to:
\begin{equation}
    \mathbf{o}_g^{i,z}=\mathbf{\alpha}_{s} \mathbf{o}_{s}.
\end{equation}


%

%


We can thus rank all $\mathbf{T}_i \in \mathbf{\Pi}$ by their descending height $\mathbf{o}_g^{i,z}$ (descending) and keep only the \textit{top-k} poses with the highest values, forming the subset
$\mathcal{G}_h \subseteq \mathbf{\Pi}$. This is defined as
\begin{equation}
\begin{aligned}
\mathcal{G}_h \;&=\; \mathcal{F}_h(\mathbf{\Pi}, \mathcal{I}, \mathcal{M}, \mathbf{\alpha}_s)
\end{aligned}
\end{equation}

\subsection{Final Set of Graspable Poses}

We combine both the \emph{IMU-based height filtering} with \emph{vision based filtering} by applying them one after the other. Typically, we first apply the inertial filter to obtain $\mathcal{G}_h \subseteq \mathbf{\Pi}$, followed by the vision filter to the get final set of graspable poses $\mathcal{G} \subseteq \mathcal{G}_h$. The final filtering function $\mathcal{F} = \mathcal{F}_v \circ \mathcal{F}_h $, given as


\begin{equation}
\begin{aligned}
\mathcal{G} \;&=\; \mathcal{F}(\mathbf{\Pi}, \mathcal{I}, \mathcal{M}, \mathbf{\alpha}_s)
\end{aligned}
\end{equation}

This final set \(\mathcal{G}\) contains filtered object poses prioritizing unobstructed objects on the top of a stack to ensure higher grasping success rates.




%% file: Dataset.tex
\section{Dataset and Ground Truth Generation}
\label{sec:dataset}

Due to the absence of publicly available stacked object datasets, we construct a comprehensive synthetic dataset with 6DoF object pose and binary graspability labels using construction bricks as our example object. The data generation and annotation process is described in the following subsections. The dataset will be made available upon acceptance of the paper.

\subsection{Synthetic Dataset Creation}  
We generated a synthetic and procedurally designed dataset using \textit{BlenderProc}~\cite{BlenderProc} designed for the BOP challenge~\cite{hodan2024bop}. 

To replicate real-world construction scenarios, we utilized a scanned 3D model of a brick preserving geometry and texture details. We created \textit{3×3} brick units and procedurally stacked them to form various configurations, including regular and irregular stacks with different levels of occlusion. Utilizing BlenderProc, we simulated real-world lighting conditions along with randomized textures for walls and floors. 

\begin{figure}[!htbp]
  \centering
  \includegraphics[width=0.48\textwidth]{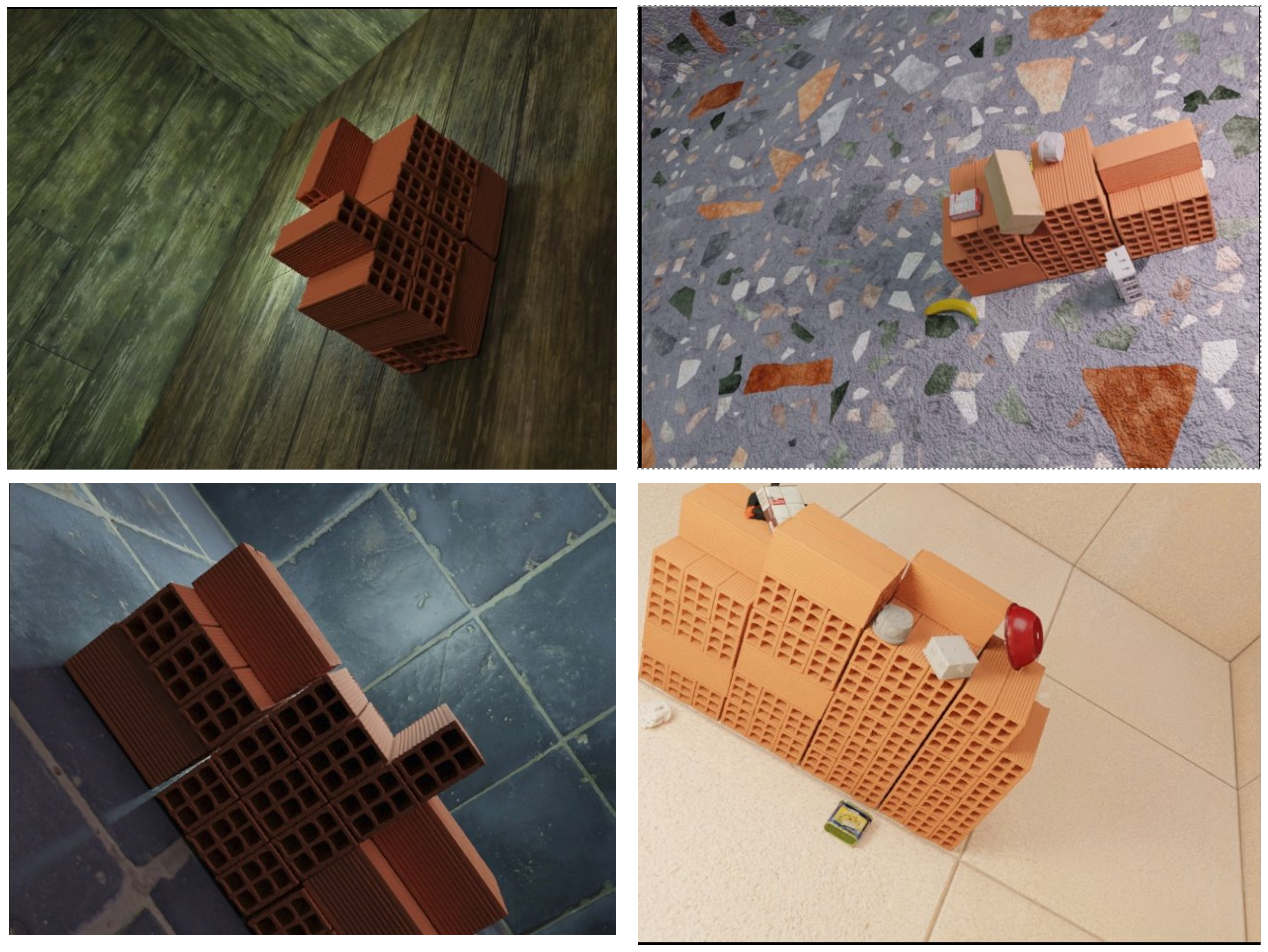}
  \caption{Example images from the synthetic dataset of stacked objects. We simulate different stack topologies with varying lighting conditions, poses, and backgrounds. Clutter objects are added in some of the images.}
  \label{fig:synthetic_images}
\end{figure}

The bricks were stacked in the middle of a room, and camera poses were sampled within a predefined spherical radius around the scene center. The process yields RGB images, depth maps, segmentation masks, and accurate 6DoF object pose annotations. Synthetic data examples from our dataset are shown in Figure \ref{fig:synthetic_images}.

\vspace{-1em}
\subsection{Ground Truth Generation}
To produce ground truth annotations of graspable objects, we employ a KD-Tree based method combined with an IoU based validation. Each object $i$ is represented by its 6DoF pose in world scene coordinates, represented as $\mathbf{T}_{wo}^i$.

%
%

Using this transformation, the eight corner points of a object are used to form a bounding box. A KD-Tree is constructed from the centers of these bounding boxes, enabling efficient nearest neighbor queries. For each object with center \( c^i \) and corresponding bounding box \( B^i \), the algorithm “moves” the center along each of the six principal directions (i.e., \( \pm X, \pm Y, \pm Z \)) by an amount equal to the object’s size in that dimension.
For each shifted bounding box, we compute the bounding boxes of its neighboring objects in the scene. We then verify if the shifted bounding box falls within the bounds of any other neighbors. This is done by checking if the shifted box has positive IoU with any of its neighbors. If no such neighbors exist, then we consider this a \textit{missing neighbor} in that direction. An object is then classified as graspable if it is missing neighbors in at least four out of the six principal directions. In cases where exactly three directions are missing, additional structural constraints are applied: (i) there must be no neighbor in the upward (\( Z+ \)) direction; (ii) at least one of the left/right (\( X+ \) or \( X- \)) directions must be missing a neighbor; and (iii) at least one of the front/back (\( Y+ \) or \( Y- \)) directions must be missing a neighbor. The detailed pseudo code of the ground truth generation process is provided in Algorithm \ref{alg:isolated_bricks}.
%


\begin{algorithm}[]
\scalebox{0.9}{%
\begin{minipage}{\linewidth}
\caption{Ground Truth Generation for graspable Bricks}
\label{alg:isolated_bricks}
\KwIn{Brick poses $\mathcal{P}$ = \{ $\mathbf{T}_{wo}^i \}_{i=1}^{n}$, 3D model points $\mathcal{M}$}
\KwOut{Set of graspable brick centers $\mathcal{Q}$}
Initialize: $\mathcal{B}\leftarrow []$, $\mathcal{C}\leftarrow []$, $\mathcal{S}\leftarrow []$\;
\For{$i=1$ \KwTo $n$}{
  \small $P^i \leftarrow \mathbf{T}_{wo}^i\,\mathcal{M} ; \quad B^i \leftarrow \text{BBoxCorners}(P^i)$ \\
  \small $c^i \leftarrow \text{mean}(B^i)$ \tcp*[l]{ \small Compute center} 
  \small $s^i \leftarrow \max(B^i)-\min(B^i)$ \tcp*[l]{ \small Compute Size}
  Append $B^i$, $c^i$, $s^i$ to $\mathcal{B}$, $\mathcal{C}$, $\mathcal{S}$\;
}
Build KD-Tree $T$ from centers $\mathcal{C}$\; 
$\mathcal{Q} \leftarrow \emptyset$\; 
\For{$i=1$ \KwTo $n$}{
  $m \leftarrow 0$\;
  \For{direction $d \in \{X,Y,Z,-X,-Y,-Z\}$}{
      \small $\tilde{c}^i \leftarrow c^i +  s^i\, \tilde{d}\;$ \tcp*[l]{\small $\tilde{d}$ is unit vector in direction d}
      $\tilde{B}^i \leftarrow B^i$ shifted by $ s^i\, \tilde{d}$\;
        $N^i \leftarrow T.\text{query}(\tilde{c}^i)$ \tcp*[l]{\small get neighbors}
      \If{for all $j \in N^i$, $j \neq i$, $IoU(\tilde{B}^i,B^j) > 0$}{
         $m \leftarrow m + 1$\;
      }
  }
  \If{$m \ge 4$}{
    Add $c^i$ to $\mathcal{Q}$\;
  }
}
\Return{$\mathcal{Q}$}
\end{minipage}%
}%
\end{algorithm}

%% file: Experiments.tex
\section{Experiments}
\label{sec:experiments}
\subsection{Evaluation Metrics}
\label{sec:metrics}
In this section we define the metrics we use to evaluate the correctness of the filtered candidate poses $\mathcal{G}$. Given an estimated candidate pose from  $\mathbf{T}_{co}^{i} \in \mathcal{G}$ and a ground truth pose $\hat{\mathbf{T}}_{co}$, we compute a score between them using the error functions Average Distance of Model Points (ADD) \cite{ADD} and Maximum Symmetry-Aware Surface Distance (MSSD) \cite{hodan2024bop}.

\vspace{\baselineskip}

\subsubsection{ADD-S} ADD quantifies how close the alignment is between a predicted pose and a ground-truth pose. Concretely, ADD measures the average pairwise distance between each model point $\mathbf{x} \in \mathcal{M}$ under these two poses. For objects with indistinguishable symmetries (e.g., rotational or reflective), the standard ADD metric may mistakenly penalize correct poses. To handle this we use the symmetric variant of ADD called \emph{ADD  for Symmetric Objects} (ADD-S) \cite{ADD} where the distance of each projected 3D point $\mathbf{x}_1$ under the predicted pose is compared to its closest symmetric point $\mathbf{x}_2$ under the ground-truth pose. The error score $\mathit{e}_{ADD-S}^{i}$ for a predicted pose $\mathbf{T}_{co}^{i}$ is defined as

 
\begin{equation}
\mathit{e}_{ADD\text{-}S}^{i}
= \frac{1}{|\mathcal{M}|} \sum_{\mathbf{x}_1 \in \mathcal{M}}
\min_{\mathbf{x}_2 \in \mathcal{M}} 
\bigl\|\,
\mathbf{T}_{co}^{i}\,\mathbf{x_1}
\;-\;
\hat{\mathbf{T}}_{co}\,\mathbf{x_2}
\bigr\|_2.
\label{eq:add_s}
\end{equation}

\begin{figure*}[!htbp]
  \centering
  \includegraphics[width=0.9\textwidth, height=3in]{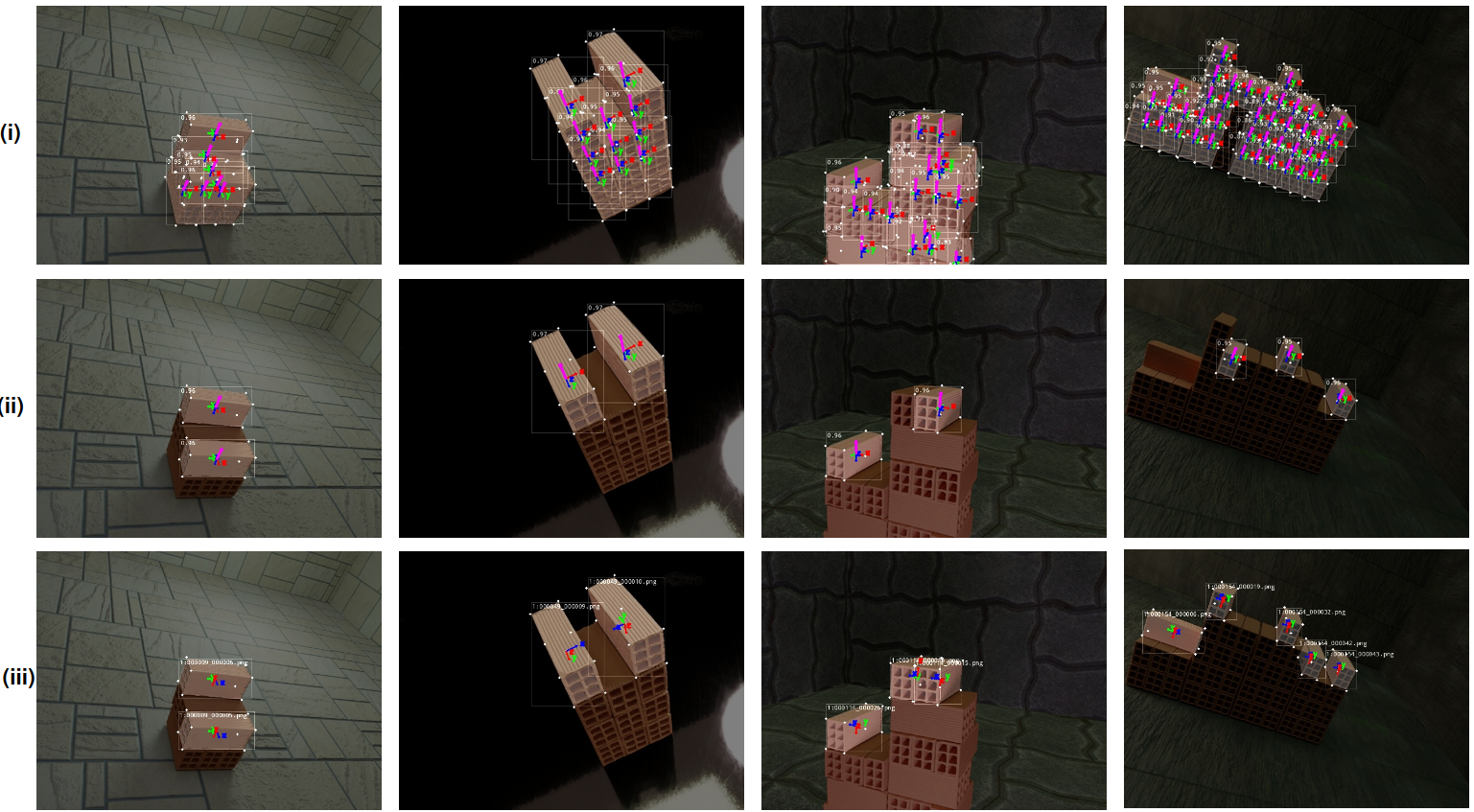}
  \caption{Qualitative results of object selection and pose estimation with our algorithm. \textit{(i)} shows all the pose estimates $\mathbf{\Pi}$ \textit{(ii)} shows the graspable candidates $\mathcal{G}$ using our filtering algorithm $\mathcal{F}$ \textit{(iii)} shows the poses from the Ground Truth.}
  \label{fig:qualitative}
\end{figure*}


\subsubsection{Maximum Symmetry-Aware Surface Distance (MSSD)}
\label{sec:mssd}

MSSD is a pose error metric that focuses on the \emph{largest} deviation between the ground-truth and predicted surfaces of an object, while accounting for global symmetries. This is especially critical in robotics, where local misalignments can lead to collisions or failed grasps. The error score $\mathit{e}_{MSSD}^{i}$ between a estimated pose $\mathbf{T}_{co}^{i}$ and the ground truth pose $\hat{\mathbf{T}}_{co}$ is defined as

\begin{equation}
\mathit{e}_{MSSD}^{i}
\;=\;
\max_{\mathbf{S} \in \mathbf{S_M}}\;
\min_{\mathbf{x} \in \mathcal{M}}\;
\Bigl\|
\mathbf{T}_{co}^{i}\,\mathbf{x}
\;-\;
\hat{\mathbf{T}}_{co}\,\mathbf{S}\,\mathbf{x}
\Bigr\|_2.
\end{equation}
where the set $\mathbf{S}$ consists of global symmetry transformation of the object model $\mathcal{M}$.
\vspace{\baselineskip}

\subsubsection{Accuracy Score}

Unlike the typical evaluation of object pose estimation methods where the pose estimates are evaluated using Average Recall (AR) scores \cite{hodan2024bop}, in our task we need to incorporate the graspability aspect of the object in to the evaluation. Thus, we focus on the filtered pose estimates set $\mathcal{G}$ and check if they are actually graspable candidates by comparing with the ground truth graspability labels. Therefore, we focus on the Precision values as the proposed metric and compute the Average Precision ($AP$). A higher $AP$ tells us how well the filtered pose candidates actually correspond to the ground truth candidates.

An estimated pose is deemed correct if its corresponding error $\mathit{e}$ satisfies $ e < (\theta \in \Theta_e),$ where  \( e \in \{ e_{\text{ADD-S}},\, e_{\text{MSSD}} \} \) and $\Theta_e$ is the set predefined thresholds for correctness. To compute Precision values, we perform optimal matching between predicted candidate poses and ground-truth candidates using the Hungarian (Kuhn-Munkres) algorithm \cite{kuhn1955hungarian} applied to a cost matrix of error scores. \emph{True Positives (TP)} are matched poses whose error scores are below $\theta$, \emph{False Positives (FP)} are predicted poses that either fail to match any ground-truth or whose error scores are above $\theta$ . The Average Precision $AP_e$ is then defined as the average of the Precision rates calculated for multiple settings of the threshold $\theta_e$. The $Precision$ and the $AP$ are defined as 
\vspace{-0.3em}
\begin{equation}
 \mathit{Precision}
 = \frac{\mathit{TP}}{\mathit{TP} + \mathit{FP}},
\quad
\mathit{AP}_e = \frac{1}{|\Theta_e|} \sum_{\theta \in \Theta_e} \mathit{Precision}(\theta)
\end{equation}

The settings for the $\Theta_e$ are similar to the BOP \cite{hodan2024bop}.

\subsection{Implementation Details}
For training and evaluating our methods, we generate a synthetic dataset with 12K images in total, consisting of 6DoF poses, modal \& amodal masks, depth and RGB images saved in BOP \cite{hodan2024bop} and COCO \cite{coco} formats. We split our dataset into 10K, 1K and 1K images for train, val and test splits respectively. In our experiments, for the filtering function $ \mathcal{F}$ we set $\epsilon_{\mathrm{vis}}$ to 0.80, as this allows only the bricks with the high visibility and thus least occlusions. Note that our approach is generic and independent of the selection of the exact detector and pose estimator.

\subsection{Benchmark evaluation}
The benchmark results for our approach are shown in Table~\ref{table:baseline}. We first run an evaluation directly on the pose estimates set $\mathbf{\Pi}$ predicted by the pose estimator. We compare the unfiltered poses with the ground truth using the metrics defined in Section \ref{sec:metrics}. Then, we evaluate with $\mathbf{top\_k = 1}$ i.e. the top brick from the filtered pose estimates is selected for evaluation. Next, we also check for $\mathbf{top\_k = 2,3}$ where 2 and 3 best bricks from the filtered pose estimates is selected for evaluation with ground truth. The results show that our method achieves good precision for selecting a single candidate for grasping and acceptable precision when we attempt to select more than one object for grasping from a single image. Qualitative examples are shown in Figure \ref{fig:qualitative}.






\begin{table}[!ht]
    \centering
    \begin{tabular}{c||c|c|c}
        \hline
        \textbf{} & $\mathbf{AP_{\text{\tiny ADD}}} \bm{^\uparrow}$ & $\mathbf{AP_{\text{\tiny MSSD}}}\bm{^\uparrow}$ & $\textbf{mAP} \bm{^\uparrow}$ \\ 
        \hline
        \hline
        \textbf{all poses}          & 0.19 & 0.16 & 0.17 \\ \hline
        \textbf{OURS (top\_k=1)}    & \textbf{0.84} & \textbf{0.74} &  \textbf{0.79}\\ \hline
        \textbf{OURS (top\_k=2)}    & 0.74 & 0.66 & 0.70 \\ \hline
        \textbf{OURS (top\_k=3)}    & 0.68 & 0.62 & 0.65 \\ \hline
    \end{tabular}
    \caption{Evaluation of precision over ADD and MSSD of filtered 6DoF pose estimates for grasping. \textit{all poses} corresponds to no filtering, \textit{top\_k} corresponds to selecting $k$ grasp candidates.}
    \label{table:baseline}
\end{table}

\subsection{Ablation Study}
Table \ref{table:ablation_1} shows our ablation experiments where we compare to baselines and to partial versions of our algorithm.  These experiments include: (i) Selecting random pose from the list of pose estimates (ii) Selecting the pose with highest detector confidence (iii) Only using the IMU information i.e., using only the height based filtering $\mathcal{G}_h$ as explained in \ref{sec:height-filter} and (iv) Only using the visual information i.e., using the vision based filter $\mathcal{G}_v$ as explained in Section \ref{sec:vision-filter}. The results clearly indicate the superiority of the combined visual-inertial method for object selection.

\begin{table}[!ht]
    \centering
    \begin{tabular}{l||c|c|c}
        \hline
        \textbf{} & $\mathbf{AP_{\text{\tiny ADD}}} \bm{^\uparrow}$ & $\mathbf{AP_{\text{\tiny MSSD}}}\bm{^\uparrow}$ & $\textbf{mAP} \bm{^\uparrow}$ \\ 
        \hline
        \hline
        \textbf{Random choice}                              & 0.18 & 0.17 & 0.17 \\ \hline
        \textbf{Detector confidence}                        & 0.64 & 0.57 & 0.60\\ \hline \hline
        \textbf{OURS - inertial only ($\mathcal{G}_h$)}     & 0.68 & 0.55 & 0.61 \\ \hline
        \textbf{OURS - vision only ($\mathcal{G}_v$)}       & 0.81 & 0.66 & 0.73 \\ \hline
        \textbf{OURS ($\mathcal{G}$) (top\_k=1)}            & \textbf{0.84} & \textbf{0.74} & \textbf{0.79} \\ \hline
    \end{tabular}
    \caption{Ablation study. We compare to two baselines for object selection, random selection from all detections and top confidence detection selection. We also evaluate our visual and inertial approaches separately and combined.}
    \label{table:ablation_1}
\end{table}

\subsection{Deployment experiments}
In addition to the evaluation on the synthetic dataset shown above, we show in the supplementary video: 1) Qualitative examples from the use of our object selection and pose estimation approach in a real video of a brick stack. 2) Videos from the deployment of the algorithm on a robot arm for brick selection and grasping from a stack. To demonstrate in a real-world scenario, we used our method at a construction site to select the best graspable bricks from a brick pallette. We deployed it on the Yaskawa six-axis HC20DTP collaborative robot arm  with a horizontal reach of 1700mm, a vertical reach of 3400mm and payload capacity of 20 kg with a custom two pronged gripper. We use an Intel Realsense d435i as our camera and IMU sensors which is connected to the flange of the robot.